\documentclass[letterpaper, 10 pt, conference]{ieeeconf}

\usepackage{cite}
\usepackage{graphicx}
\usepackage{framed}
\usepackage{subfig}
\usepackage{indentfirst}
\usepackage[]{algorithm2e}

\usepackage{amsmath, amssymb,amsthm}
\usepackage[font=small]{caption}

\IEEEoverridecommandlockouts                
\title{\LARGE \bf 
Towards a Fatality-Aware Benchmark of Probabilistic Reaction Prediction in Highly Interactive Driving Scenarios}
\author{Wei Zhan$^{1}$, Liting Sun$^{1}$, Yeping Hu$^{1}$, Jiachen Li$^{1}$, 
	and Masayoshi Tomizuka$^{1}$
\thanks{*This work was partially supported by the international Chair Drive for All, Foundation MINES ParisTech, and was partially supported by Berkeley DeepDrive (BDD).}
\thanks{$^{1}$W. Zhan, L. Sun, Y. Hu, J. Li, and M. Tomizuka are with the Department of Mechanical Engineering, University of California, Berkeley, CA 94720 USA (e-mail: \tt\small wzhan, litingsun, yeping\_hu, jiachen\_li, tomizuka@berkeley.edu). }
}

\begin{document}
\maketitle

\begin{abstract}
In order to achieve safe and high-quality decision-making and motion planning, autonomous vehicles should be able to generate accurate probabilistic predictions for uncertain behavior of other road users. Moreover, reactive predictions are necessary in highly interactive driving scenarios to answer ``what if I take this action in the future" for autonomous vehicles. Many recently proposed methods based on probabilistic graphical models (PGM), neural networks (NN) and inverse reinforcement learning (IRL) have great potential to solve the problem. However, there is no existing unified framework to homogenize the problem formulation, representation simplification, and evaluation metric for those methods. In this paper, we formulate a probabilistic reaction prediction problem, and reveal the relationship between reaction and situation prediction problems. We employ prototype trajectories with designated motion patterns other than ``intention" to homogenize the representation so that probabilities corresponding to each trajectory generated by different methods can be evaluated. We also discuss the reasons why ``intention" is not suitable to serve as a motion indicator in highly interactive scenarios. We propose to use Brier score as the baseline metric for evaluation. In order to reveal the fatality of the consequences when the predictions are adopted by decision-making and planning, we propose a fatality-aware metric, which is a weighted Brier score based on the criticality of the trajectory pairs of the interacting entities. Conservatism and non-defensiveness are defined from the weighted Brier score to indicate the consequences caused by inaccurate predictions. Modified methods based on PGM, NN and IRL are provided to generate probabilistic reaction predictions in an exemplar scenario of nudging from a highway ramp. The results are evaluated by the baseline and proposed metrics to construct a mini benchmark. Analysis on the properties of each method is also provided by comparing the baseline and proposed metric scores.
\end{abstract}

\section{Introduction\label{sec: introduction}}


Accurate prediction of the future motions of surrounding entities is a prerequisite for autonomous vehicles to make decisions and plan motions under uncertainties \cite{hubmann2017decision}\cite{sefati2017towards}\cite{chen_continuous_2018} that are safe with high driving quality. Probabilistic prediction is necessary since the human behavior is full of uncertainties. The accuracy of generated prediction probabilities can significantly impact the safety and driving quality of autonomous vehicles. As was stated in \cite{zhan_non-conservatively_2016}, a desirable driving strategy of autonomous vehicles should be defensive to real threats, but not conservative to threats of low probability. A fatal accident may happen if the prediction algorithm ignores a real threat by mistaking its probability as zero, which makes the driving strategy non-defensive. On the other hand, the decisions and motions of autonomous vehicles can be very conservative if the prediction algorithm overestimates the probability of a threat and generates false alarms. Therefore, accurate probabilistic prediction is a key building block for safe and high-quality autonomous driving.


Many probabilistic prediction methods have been proposed based on neural networks (NN) \cite{phillips_generalizable_2017}\cite{morton_analysis_2017}\cite{hu2018SIMP}, as well as probabilistic graphical models (PGM) \cite{dong_intention_2017} such as particle filter \cite{hoermann_probabilistic_2017}\cite{li2018tracking} and Bayes net \cite{schreier_integrated_2016}. All the aforementioned literatures formulated the prediction problem to predict the distribution of future motions of an entity given the historical motions of relevant participants in the scene. However, solving such a problem cannot provide sufficient and accurate predictions for highly interactive driving scenarios. When several entities are closely interacting with each other in a specific scenario, the future motion of the host autonomous vehicle can significantly impact the motion of its surrounding entities. Therefore, like human drivers, autonomous vehicles should always ask ``what if I take this action" during interactions. Recently, prediction methods based on inverse reinforcement learning (IRL) \cite{sadigh2016planning} were proposed to tackle reaction prediction problems. However, only optimal motions were provided in IRL, which makes it deterministic. Therefore, the main stream paradigms for prediction, such as NN, PGM and IRL, should be modified to solve probabilistic reaction prediction problems for highly interactive driving scenarios. 

Moreover, as was discussed in \cite{zhan_2018}, the problem formulation and motion representation of different methods should be homogenized within a unified framework so that they can be evaluated. The distribution to be approximated for probabilistic reaction prediction should be explicitly defined. Moreover, for methods except for IRL, approximating the (situational) joint distribution \cite{li2018hmm}\cite{hu2018vae} \cite{klingelschmitt_situation_2016}\cite{lefevre_intention-aware_2013} of the motions of several entities is typically much easier than approximating the reaction distribution. Therefore, we should also provide the transformation between situation and reaction predictions. Since it is intractable to approximate the distribution of continuous random variables (long-term 2D trajectories) with high dimension, we also need to simplify the representation of the motions to discrete motion patterns. The simplified representation should also be homogenized in spatiotemporal domain to construct a unified framework for different methods.



Appropriate evaluation metrics should be selected to measure the performance of probabilistic prediction algorithms. Metrics such as root mean square error (RMSE) \cite{morton_analysis_2017}\cite{wheeler_analysis_2016}, likelihood \cite{phillips_generalizable_2017}\cite{wheeler_analysis_2016} and Kullback-Leibler (KL) divergence \cite{morton_analysis_2017}\cite{wheeler_analysis_2016} were employed in existing works to evaluate the performance of probabilistic prediction methods and algorithms. Each of the metrics has its own limitations which may lead to difficulties or misinterpretations to measure the performance of probabilistic prediction. We need to choose a proper metric to measure the performance of prediction algorithms to approximate the data distribution.


The purpose of probabilistic prediction is not limited to approximate the data distribution. The prediction results are used online for decision-making and planning modules. Safe and high-quality motions are expected by adopting the prediction outputs. Therefore, the evaluation metric should reveal the decision consequence due to the inaccurate prediction, such as how non-defensive or conservative the planned motion would be, and what is the fatality of the consequence. A fatality-aware metric for prediction is expected, which has not been addressed in existing works. Also, when evaluating predicted motions of surrounding entities, we should also take into account prior knowledge such as vehicle kinematics and dynamics, as well as rare collision in the real world. In fact, satisfying feasibility and safety requirements is hard to achieve for many existing methods.

In \cite{zhan_2018}, the under-explored aspects of probabilistic prediction were highlighted with a comprehensive review on problem formulation, representation simplification and evaluation metric. In this paper, we provide a preliminary solution for the problems pointed out in \cite{zhan_2018}, and implement the ideas with evaluation results as a mini benchmark. The main contribution of this paper is to propose a fatality-aware evaluation metric for probabilistic reaction prediction in extremely challenging driving scenarios with interaction. The proposed metric can reveal the fatality of prediction errors by considering the criticality of the corresponding motion pair. Moreover, we construct a unified framework with homogenized problem formulation and motion representation, which can evaluate different types of methods such as PGM, NN and IRL. We implement these methods in highly interactive ramp merging scenarios, and evaluate the prediction performances with the proposed metrics with analysis.

\section{Problem Formulation}

In this section, the problem to be solved by probabilistic reaction prediction is formulated. Suppose $q^i$ and $\hat{q}^i$ represent the historical and (predicted) future motions of the $i$th entity, respectively. The host vehicle corresponds to $i=0$, and the predicted entity corresponds to $i=1$. Suppose there are $N$ entities to be considered around the host vehicle in the scene at the current time step. Then the original problem which is typically tackled in existing literatures is to obtain desirable models to approximate the conditional probability density function (PDF)
\begin{eqnarray}
p ( \hat{q}^1 | q^{0:N}).
\label{eq1}
\end{eqnarray}

However, what is required by the prediction in highly interactive scenarios is far beyond perfectly solving the original problem. When the driving scenarios are highly interactive, the host autonomous vehicle cares about not just the future motion of the predicted entity condition on the historical motions of all relevant entities. The reaction of the predicted entity given different future motions of the host vehicle should also be taken into account. Then the conditional PDF to be approximated for reaction prediction can be written as 
\begin{eqnarray}
p ( \hat{q}^1 | q^{0:N}, \hat{q}^0),
\label{eq2}
\end{eqnarray}
where $\hat{q}^0$ is the future motion of the host autonomous vehicle.

From the perspective of designing learning algorithms, approximating the reaction distribution (\ref{eq2}) is much harder than approximating a situational joint distribution of the interaction pair. A situation distribution to be approximated can be written as
\begin{eqnarray}
p ( \hat{q}^0, \hat{q}^1 | q^{0:N} ),
\label{eq3}
\end{eqnarray}
and it is easy to transform (\ref{eq3}) to (\ref{eq2}) since
\begin{eqnarray*}
p ( \hat{q}^1 | q^{0:N}, \hat{q}^0)=\frac{p ( \hat{q}^0, \hat{q}^1, q^{0:N} )}{p ( q^{0:N} ) p ( \hat{q}^0 | q^{0:N} )} = \frac{p ( \hat{q}^0, \hat{q}^1 | q^{0:N} )}{\int_{\hat{q}^1}^{} p ( \hat{q}^0, \hat{q}^1 | q^{0:N} )}.
\end{eqnarray*}
It means that we can design learning algorithms to learn how to predict the joint distribution of the motions of the predicted entity and the host vehicle. Then we can transform it into reaction predictions.

\section{Representation Simplification\label{sec: representation}}

It is intractable to approximate the distribution of long-term trajectories, which are continuous random vectors of motions with high dimensions. Therefore, the representation of future motions is typically simplified by using discrete indicators. In this section, we first clarify the distinction between desires in human mind and executed motion patterns. The clarification explains why ``intention" is not suitable to be the simplification indicator. Then we illustrate how to construct the spatiotemporal representation of potential motion patterns. 

\subsection{Clarification of desire and pattern}

	\begin{figure*}[t!]
	\begin{center}
		\includegraphics[width=17.5cm]{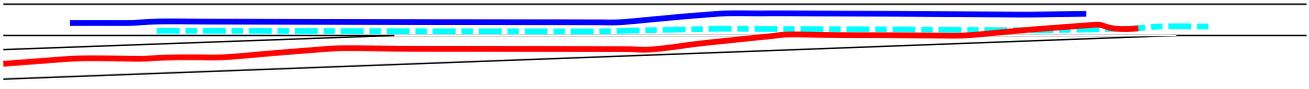}
		\caption{Trajectories of the merging (red), target (blue) and front (cyan) vehicles in a ramp merging scenario. }
		\label{fig: case14traj}
	\end{center}
	\end{figure*}

	\begin{figure}[t!]
	\begin{center}
		\includegraphics[width=8.5cm]{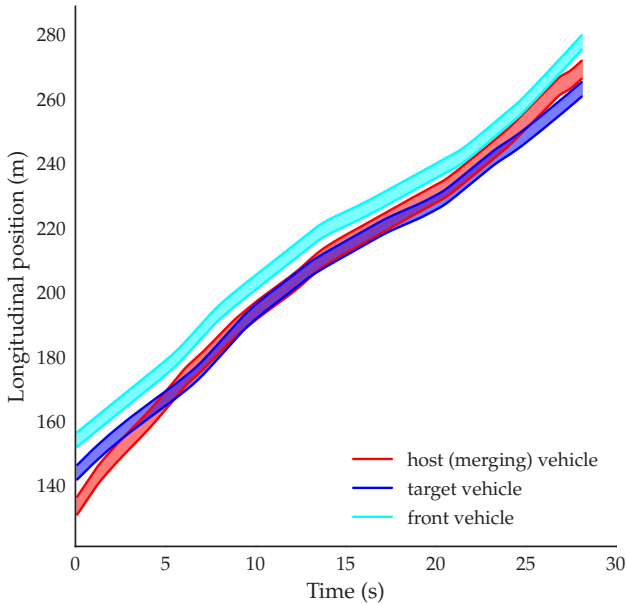}
		\caption{longitudinal positions over time of the merging (red), target (blue) and front (cyan) vehicles in a ramp merging scenario.}
		\label{fig: case14station}
	\end{center}
	\end{figure}

``Intention" is typically used as an indicator to simplify the motion representation. However, it often denotes either the desire in the human mind or the executed motion pattern, which is not explicit. As was pointed out in \cite{zhan_2018}, it is impossible to obtain the strict ground truth of the desire in the human mind, especially in highly interactive driving scenarios. We will use the following example of ramp merging on I-80 in NGSIM dataset \cite{kim2005machine} to explain.

Figure \ref{fig: case14traj} illustrates the trajectories of a ramp merging scene with a preceding and a following target vehicle on the target lane, as well as a merging vehicle on the ramp. The merging vehicle tried to merge into the gap between the preceding and target vehicle, but the target vehicle refused to yield to create a gap. The merging and target vehicles were driven side by side in parallel for a long period. After an adversarial interaction procedure for around 20 seconds, the target vehicle enlarged the gap to enable the merging vehicle to cut in successfully. The procedure in spatiotemporal domain can be found in Figure \ref{fig: case14station} with the longitudinal positions of each vehicle at each time step.
	
In fact, it is impossible to obtain the ground truth of the desires in the human mind at each time step. For instance, the gap became larger during $5$-$7$ s and went back to a small one during $7$-$9$ s. The desire in the mind of the target vehicle driver on whether to yield the merging vehicle can change from time to time. Also, it is unclear whether the driver of the merging vehicle hesitated or even decided to give up and merge into the gap behind the target vehicle around $20$-$22$ s since it was approaching the end of the ramp.

In conclusion, we can only observe the executed motion (pattern) in the dataset as the ground truth for highly interactive scenarios. ``Intention" representing desires in human mind is not an appropriate indicator to simplify the representation since there is no ground truth to compare. Therefore, we use executed motion patterns in this paper to simplify the representation.

\subsection{Motion pattern in a spatiotemporal domain}

As was reviewed and summarized in \cite{zhan_2018}, motion patterns can be categorized hierarchically into route, pass-yield and subtle patterns in various kinds of scenarios. In different driving scenarios with different situations of the entities, the choice of the patterns at each level can be completely different. In general, we suppose there are $M$ possible motion patterns for the predicted entity in a specific highly interactive driving scenario. $\hat{s}^i_j$ represents the $j$th future motion pattern of the $i$th entity. Then the distribution to be approximated can be written as 
\begin{eqnarray}
P ( \hat{s}_j^1 | q^{0:N}, \hat{s}^0), \ j=1,2,...,M.
\label{eq4}
\end{eqnarray}

In \cite{zhan_2018}, two spatiotemporal representations were summarized, namely, prototype trajectory and reachable set, which can incorporate the generated motion patterns. In this paper, we adopt prototype trajectory as the spatiotemporal representation. The main purpose is to make it convenient for the methods and algorithms in Section \ref{sec: method} to obtain the corresponding probabilities since the outputs of those algorithms are trajectories. Suppose the generated prototype trajectory corresponding to motion pattern $\hat{s}_j^i$ is expressed as $\hat{q}_j^i$. Then we can use the normalized probabilities of
\begin{eqnarray}
p ( \hat{q}_j^1 | q^{0:N}, \hat{q}^0), \ j=1,2,...,M.
\label{eq10}
\end{eqnarray}
obtained from the distribution generated by learning methods to calculate (\ref{eq4}).


For a given motion pattern, we modify the motion planning approach proposed in \cite{zhan_spatially-partitioned_2017} to generate trajectories. The framework proposed in \cite{zhan_spatially-partitioned_2017} can deal with various kinds of driving scenarios, and generate smooth, feasible and collision-free (if necessary) trajectories. The computation is extremely fast since only simple A* search and quadratic programming (QP) are involved. The approach is suitable for both online prediction generation, as well as offline evaluation of trajectories for which a large set of predicted trajectories need to be generated.

\section{Methodologies\label{sec: method}}

In this section, the methodologies to generate probabilistic reaction prediction results are briefly introduced, including hidden Markov model (HMM), mixture density network (MDN) and inverse reinforcement learning (IRL).

\subsection{Hidden Markov model (HMM)}

A hierarchical motion prediction framework is employed. It is essentially a cascade of a situation inference module based on a group of hidden Markov model (HMM), and a motion prediction module based on a group of Gaussian mixture model (GMM) corresponding to each interaction outcome. The labeled trajectories of each potential situation are used to train a HMM individually with the Baum-Welch algorithm, which is a variant of Expectation-Maximization (EM) algorithm. The GMM is used to obtain the conditional distribution of the actions of multiple interactive entities given the current state information.

At the inference stage, given a sequence of historical motion, we can obtain the likelihood of the observation sequences for each HMM by the forward algorithm. Then the likelihood values are normalized to obtain the posterior probability of each situation, which can be written as $p^{k'}$, where $k'=1,...,K$, and $K$ is the number of possible situations according to the combination of pass-yield motion patterns \cite{zhan_2018}. For each GMM, we can obtain the probabilistic density for the $j$th prototype trajectory, denoted as $f^k_j$.
Then the probability of the $j$th prototype trajectory can be obtained by
\begin{equation}
P(\hat{s}^1_j|q^{0:N},\hat{s}^0) = \frac{ \displaystyle{\sum_{k'=1}^{K}} p^{k'} f^{k'}_j}{ \displaystyle{\sum_{j'=1}^M \sum_{k'=1}^K} p^{k'} f^{k'}_{j'}}.
\end{equation}

\subsection{Mixture density network (MDN)}

We use the mixture density network (MDN) \cite{bishop1994mixture} to obtain the joint PDF of the host vehicle and other entities. Instead of learning a single output value using neural networks, MDN is capable of predicting an entire probability distribution for the output using a Gaussian Mixture Model (GMM). Given a set of input states and output actions of multiple traffic participants, MDN can generate necessary parameters to formulate the conditional probability of actions given states. 

Given a sequence of prototype trajectory, the performed actions by the vehicle between each state can be obtained. Then at each time step, we can forward the current state into the MDN network and use the conditional distribution to get the likelihood of the action at the given state. For each motion pattern and its corresponding trajectory sequence, we can then multiply the obtained likelihood over the entire horizon and perform a normalization to get the posterior probability for each situation.

Then the approximated distribution for the $j$th future motion pattern can be formulated as:
\begin{equation}
P(\hat{s}_j^1|q^{0:N},\hat{s}^0) = \dfrac{\displaystyle{\sum_{n=1}^{N_m}} w_n\phi_n(\hat{q}_j^1|q^{0:N},\hat{q}^0)}{\displaystyle{\sum_{j'=1}^{M}\sum_{n=1}^{N_m}} w_n\phi_n(\hat{q}_{j'}^1|q^{0:N},\hat{q}^0)},
\end{equation}
where $N_m$ denotes the total number of mixture components, $w_n$ is the mixing coefficient and $\phi(\hat{q} | q)$ is the kernel function.

\subsection{Inverse reinforcement learning (IRL)}

Inverse reinforcement learning allows us to learn the cost functions of human by observing their behavior. We assume that all predicted agents are rational, and their cost function along a motion trajectory $q$ can be linearly parametrized as $C(\theta, q){=}\theta^T\mathbf{f}(q)$ where $\mathbf{f}(q)$ are features. We also assume that trajectories with lower cost are exponentially more probable based on the principle of maximum entropy \cite{ziebart2008maximum}:
\begin{equation}
\label{eq:continuous_irl}
P({q}|\theta)\propto e^{-C(\theta, q)}
\end{equation}
In the training phase, the goal of IRL is to find the optimal $\theta^*$ that best explains the observed demonstrations in terms of a set of selected features $\mathbf{f}(q)$. Mathematically, we need to solve the following optimization problem:
\begin{eqnarray}
\label{eq:likelihood_Xi}
\theta^*(\mathcal{Q})=\arg\max_{\theta}\prod_{{\tilde{q}}\in\mathcal{Q}}\dfrac{e^{-C(\theta, q)}}{\int e^{-C(\theta, \tilde{q})}d\tilde{q}}.
\end{eqnarray}
where $\mathcal{Q}$ represents the set of demonstrated trajectories.

With $\theta^*$, an exponential distribution family is established to approximate the distribution of future trajectories. Different from approaches based on probabilistic graphical models and neural networks, IRL directly generates the conditional probability defined in (\ref{eq2}) instead of generating the joint distribution over trajectories of the two interacting entities.

In the test phase, given a set of sampled motion patterns $\mathcal{S}$, we can evaluate the normalized probability of each motion pattern  $\hat{s}_j^1{\in}\mathcal{S}$ via:
\begin{equation}
\label{eq:normalization}
P(\hat{s}_j^1|q^{0:N}, \hat{s}^{0}) = \dfrac{e^{-C(\theta, \hat{s}_j^1|q^{0:N}, \hat{s}^{0})}}{\displaystyle{\sum_{\tilde{s}^1{\in}\mathcal{S}}}e^{-C(\theta, \tilde{s}^1|q^{0:N}, \hat{s}^{0})}}。
\end{equation}
More details on IRL-based probabilistic prediction can be found in \cite{sun2018irlprediction}.

\section{Evaluation Metric\label{sec: metric}}

In this section, we address three aspects to obtain appropriate evaluation metrics. The first is on whether the predicted motions satisfy safety and feasibility requirements based on prior knowledge. Next is to select an appropriate baseline metric. Finally, we propose the fatality-aware metric based on the baseline.

\subsection{Prior knowledge}

Based on our prior knowledge, the predicted motion of vehicles should at least satisfy a simple kinematic model, and collisions should be extremely rare according to the statistics of real-world driving. In other words, safety and feasibility should also be checked when evaluating the prediction performance. However, it is a difficult task for algorithms based on pure neural networks or probabilistic graphical models to satisfy feasibility and safety constraints. 

Since we are using the prototype trajectories generated in Section \ref{sec: representation} to represent possible motion patterns, it is relatively easy to make the generated trajectories satisfy the requirements on safety and feasibility. It can alleviate the requirements on safety and feasibility for learning-based models, and additional verifications are not necessary.

\subsection{Baseline metric}

Existing works typically use metrics such as area under the curve (AUC), likelihood, root mean square error (RMSE), and Kullback-Leibler (KL) divergence to evaluate probabilistic predictions. In this subsection, we briefly discuss the deficiencies of each metric when it is employed for evaluation of probabilistic predictions.

AUC is a metric which is typically used for binary classification, which is not inherently designed to reveal the accuracy of the probabilistic distributions. Likelihood sufficiently measures accuracy of the predicted probabilities or distribution for the ground truth data points. However, it is not possible to indicate how bad the prediction is if high density is generated for motions with low or zero probability. 

RMSE measures the error between the ground truth and sampled trajectories from the predicted distribution in Euclidean space. However, the Euclidean distance can be very small between collision-free, critical and colliding trajectories, or between feasible and infeasible trajectories. A small perturbation of the trajectory in Euclidean space can make it completely different on whether or not the trajectory is collision-free or feasible. Also, RSME fails to reveal the approximation performance for multimodal data \cite{rhinehart2018r2p2}.


KL divergence requires the description of the ground truth distribution, which is extremely hard to obtain in a high dimensional space. Estimation or approximation of the distribution of the test data may not be executable with limited data points in the high dimensional space of the motions. 

Therefore, we need an appropriate metric without the aforementioned deficiencies. Brier score \cite{brier1950verification} is a metric measuring the accuracy of probabilistic predictions, which is widely used in research fields requiring evaluation of probabilistic predictions, such as weather forecast. It can evaluate the prediction performance directly from the ground truth data points without estimated distribution of the test set. Also, the score can both reward high probability for ground truth patterns and penalize overestimation of other patterns. By properly generating motion patterns, the score can also avoid the problem introduced by using Euclidean space.
 
Suppose $\hat{s}_g^0(k,T_h)$ and $\hat{s}_g^1(k,T_h)$ are the ground truth motion pattern of the host and predicted vehicle future motion with preview horizon $T_h$ for the $k$th sample in the test set of data. The total number of samples is $N_s$. We define 
\begin{align*}
P_j (k,T_h) = P ( \hat{s}_j^1(k,T_h) | q^{0:N}, \hat{s}^0_g(k,T_h)).
\end{align*} 
We also define
\begin{align*}
O_j (k,T_h) = O ( \hat{s}_j^1(k,T_h) | q^{0:N}, \hat{s}^0_g(k,T_h))
\end{align*}
to represent the actual outcome ($0$ or $1$) on whether the $j$th motion pattern corresponds to the ground truth. According to the definition of Brier score, the baseline metric for probabilistic prediction can be written as
\begin{align}
\mathcal{B}(T_h)= & \frac{1}{N_s M} \displaystyle{\sum _{ k=1 }^{ N_s }\sum _{ j=1 }^{ M }}  [ P_j (k,T_h) - O_j (k,T_h)]^2 .
\label{eq5}
\end{align}

\subsection{Fatality-aware metric}

The baseline metric (\ref{eq5}) equally weights each probability error $(P-O)^2$. However, each error may have different impact to the prediction accuracy due to the difference of the criticality of each motion pairs.


Suppose Cr$_{j_1,j_0}(k,T_h)$ denotes a score of the criticality for the motion pair of the predicted entity $\hat{s}_{j_1}^1(k,T_h)$ and host vehicle $\hat{s}_{j_0}^0(k,T_h)$. It can be the inverse of time-to-collision (TTC) or other scores which represents how critical motion pair is for a potential collision. Since only the ground truth motion pattern of the host vehicle is used for evaluation, we use Cr$_{j}(k,T_h)$ for the motion pair of the predicted entity $\hat{s}_{j}^1(k,T_h)$ and host vehicle ground truth $\hat{s}_{g}^0(k,T_h)$

	\begin{figure}[t!]
	\begin{center}
		\includegraphics[width=8.5cm]{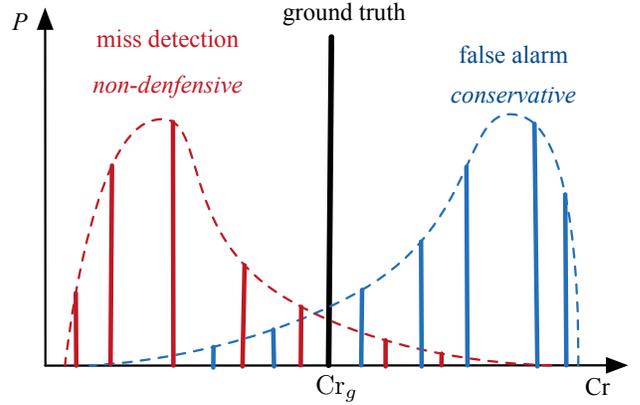}
		\caption{An intuitive illustration of conservatism and non-defensiveness by using two distributions with similar probability outputs for ground truth and similar Brier scores.}
		\label{fig: distribution}
	\end{center}
	\end{figure}

Then we can sort the predicted motion patterns according to the criticality from low to high as $\left\lbrace  \hat{s}_{1:m}^1, \hat{s}_g^1, \hat{s}_{m+2:M}^1 \right\rbrace $. Motion patterns $ \hat{s}_{1:m}^1$ are less aggressive than the ground truth, while $ \hat{s}_{m+2:M}^1$ are more aggressive. Cr$_g(k,T_h)$ is the criticality of motion pair $\hat{s}^0_g(k,T_h)$ and $\hat{s}^1_g(k,T_h)$. 

An intuitive illustration is provided in Figure \ref{fig: distribution} to explain how we can use the values of criticality to achieve better evaluation of probabilistic predictions. There are two predicted distributions with similar Brier scores and similar probability outputs for ground truth (similar likelihood). The distribution colored blue assigns high probabilities to motions which are much more critical than the ground truth. Such inaccurate predictions overestimate the threats and generate false alarms. It can lead to conservative decisions and motions of the host autonomous vehicle. By contrast, the distribution colored red provides high probabilities to motions which are much less critical than the ground truth. Such inaccurate predictions underestimate the real threat (ground truth) and generate miss detections. It can lead to non-defensive decisions and motions of the host autonomous vehicle, and may cause fatal accidents.

Therefore, the inherent distinction of the consequences (conservative and non-defensive) by adopting the predicted distribution should be revealed in the evaluation metric. Also, the more the criticality deviates from the ground truth, the more penalty should be received by the corresponding predicted probability. We propose to separate the baseline metric (\ref{eq5}) into three parts to achieve such purpose.

The first part contains the prediction errors of ground truth reaction. which can be written as
\begin{align}
\mathcal{G}(T_h)= \displaystyle{\sum _{ k=1 }^{ N_s }} \frac{1}{N_s M}  [ P_j (k,T_h) - 1]^2 .
\label{eq6}
\end{align}
The weights remain the same as the baseline metric (\ref{eq5}).

For all motion patterns which are more aggressive than the ground truth, the probability errors $(P_{m+2:M} - 0)^2$ correspond to false alarms on more dangerous reactions than what really happened. In other words, the prediction algorithm overestimates the aggressiveness of the reaction of the others. It makes the decision of the host autonomous vehicle conservative to false threats. Therefore, we denote a score to measure \textit{conservatism} of the prediction algorithm based on the baseline metric (\ref{eq5}), that is
\begin{align}
\mathcal{C}(T_h)= & \displaystyle{\sum _{ k=1 }^{ N_s }\sum _{ j=m_k +2 }^{ M }}
\frac{\text{Cr}_j (k,T_h) -\text{Cr}_g (k,T_h)}{S}  P_j (k,T_h)^2 ,
\label{eq7}
\end{align}
in which $S$ is the summation of all weights other than those for the ground truth. We use $S$ to normalize the weights.

For all motion patterns which are less aggressive than the ground truth, the probability errors $(P_{1:m} -0)^2$ correspond to miss detections. In other words, the prediction algorithm fails to predict a more aggressive reaction, which is the ground truth. It makes the decision of the host autonomous vehicle less defensive to real threats. Therefore, we denote a score to measure \textit{non-defensiveness} of the prediction algorithm based on the baseline metric (\ref{eq5}), that is
\begin{align}
\mathcal{D}(T_h)= & \displaystyle{\sum _{ k=1 }^{ N_s }\sum _{ j=1 }^{ m_k }}
\frac{\text{Cr}_g (k,T_h) - \text{Cr}_j (k,T_h)}{S}  P_j (k,T_h)^2 ,
\label{eq8}
\end{align}

Then the fatality-aware weighted metric can be written as follows, which contains the aforementioned three aspects.
\begin{align}
\mathcal{B}_c(T_h)= \mathcal{D}(T_h) + \mathcal{G}(T_h) + \mathcal{C}(T_h).
\label{eq9}
\end{align}

%
%

\section{Case Study}

In this section, the metrics in Section \ref{sec: metric} are employed to evaluate the performances of the algorithms in Section \ref{sec: method}. We use an exemplar scenario to provide a mini benchmark for the three methodologies. Highway ramp merging is a highly interactive driving scenario. It can be extremely challenging when the traffic flow is relatively slow, where the merging vehicles have to nudge into a gap on the target lane. The gaps are often very small so that the merging vehicle have to interact with a target vehicle to force it to enlarge the gap. The merging may fail and the merging vehicle have to resort to the next gap behind the target vehicle. 

The ramp merging cases in NGSIM dataset were used for the training and evaluation of the algorithms. The cases were manually selected to be highly interactive ones. In each case, we only chose the frames in which the merging vehicle has not merge into the target lane successfully and it is still interacting with the target vehicle. $7102$ data samples were used for training, and $708$ data samples were used for evaluation. 

\begin{figure}[t!]
	\begin{center}
		\includegraphics[width=8.5cm]{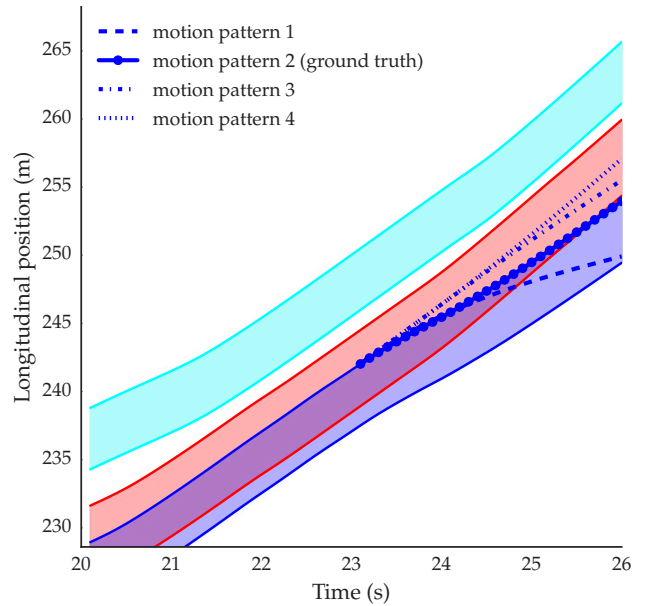}
		\caption{An illustrative example of prototype trajectories represented by longitudinal positions over time.}
		\label{fig: prototraj}
	\end{center}
	\end{figure}

The merging and target vehicle was treated as the host and predicted vehicle, respectively. The preview horizon $T_h = 3$ s. The number of motion patterns of the reaction of the target vehicle was set as $M=4$. An illustrative example of prototype trajectories represented by longitudinal positions over time was provided in Figure \ref{fig: prototraj}. It was a segment in the ramp merging case shown in Figure \ref{fig: case14station}. The current time step is $23$ s. The target actually started to yield the merging vehicle. Motion pattern 1 was less aggressive than the ground truth motion pattern. Motion patterns 3 and 4 were trying to keep the small gap, which were more aggressive than the ground truth.

The criticality score Cr$_j$ was defined as the inverse of time-to-collision (TTC) for the front end of the target vehicle to hit the (potential) merging point of the merging vehicle \cite{zhao_accelerated_2017}. Merging point was defined as the longitudinal position of the rear end of the merging vehicle when its (potential) vehicle body overlaps with the potential path of the target vehicle.

The performance scores according to the aforementioned metrics are shown in Table \ref{table: 1} for hidden Markov model (HMM), mixture density networks (MDN) and inverse reinforcement learning (IRL). If we use the baseline Brier score as the metric, HMM and MDN outperformed IRL since the score of IRL was much higher than those of the other two. However, if we partition the score $\mathcal{G}$ for the error at the ground truth data points, we can find that $\mathcal{G}$ had great impact to raise $\mathcal{B}$. It revealed that methods approximating the distribution directly trained with log likelihood, such as HMM and MDN, can easily outperform IRL if we only care about assigning high probabilities at ground truth data points. 

For \textit{conservatism} $\mathcal{C}$ and \textit{non-defensiveness} $\mathcal{D}$, the scores of IRL were much lower than those of the other two methods. It demonstrated that IRL tended to produce relatively high probabilities for prototype trajectories with similar criticality as the ground truth. The reason may come from the nature of IRL to approximate the reward/cost function other than the data distribution, which makes the motions generated by IRL more interpretable. Such properties can help the host autonomous vehicle to avoid conservative behaviors by reducing the probabilities of potential reactions which are much more critical than the ground truth. More importantly, the autonomous vehicle can behave defensively to avoid underestimating real threat. The advantage of such property was also reflected in the criticality-aware metric $\mathcal{B}_c$. The score of IRL became lower than those of the other two.

\begin{table}[tbp]
	\caption{Performance scores of each method}\centering{}%
	\begin{tabular}{c|ccc}
		& HMM & MDN & IRL \\ \hline
		$\mathcal{B}$ & $0.1403$ & $0.1099$ & $0.1821$ \\
		$\mathcal{G}$ & $0.0710$ & $0.0564$ & $0.1117$ \\
		$\mathcal{C}$ & $0.0476$ & $0.0493$ & $0.0178$ \\
		$\mathcal{D}$ & $0.1365$ & $0.1303$ & $0.0698$ \\
		$\mathcal{B}_c$ & $0.2551$ & $0.2361$ & $0.2053$%
	\end{tabular}%
\label{table: 1}
\end{table}

Note that what we discussed in this Section analyzed the specific methods we implemented. It does not necessarily conclude that one paradigm or one type of methodologies is better than others in those aspects. By properly modifying the methods, tuning parameters or redesigning the framework, better performances can be achieved.

\section{Conclusion\label{sec: conclusion}}
In this paper we proposed a unified framework with a fatality-aware metric to evaluate the performance of probabilistic reaction prediction in highly interactive driving scenarios. Three methods based on probabilistic graphical model (PGM), neural network (NN) and inverse reinforcement learning (IRL) were modified with homogenized problem formulation and representation simplification. By using prototype trajectories with designated motion patterns as the simplified representation, the requirements on collision avoidance and feasibility can be satisfied. We employed Brier score as the baseline metric to overcome the deficiencies of the existing metrics. We proposed a weighted Brier score based on the criticality of the interactive motion pairs. The proposed evaluation metric emphasized the fatality of the consequences when corresponding predictions are adopted. Conservatism and non-defensiveness were also defined based on the proposed metric for analyzing the performance of the prediction algorithms. Analysis on the implemented methods was provided by comparing the baseline and proposed metric scores of each method. 

\bibliography{benchmark}

\end{document}